\documentclass{article} % For LaTeX2e
\usepackage{nips15submit_e,times}
\usepackage{hyperref}
\usepackage{graphicx}
\usepackage{url}
\usepackage{cite}
\usepackage{amsmath}
\usepackage{subcaption}
\usepackage{amssymb}
\usepackage{float}
\usepackage{booktabs}
\usepackage{hyperref}
\usepackage{cleveref}
\usepackage{multirow}
\title{Video Moment Localization using Object Evidence and Reverse Captioning}
\author{Madhawa Vidanapathirana, Supriya Pandhre, Sonia Raychaudhuri, Anjali Khurana\\
Simon Fraser University\\
\texttt{\{mvidanap,spandhre,sraychau,anjali\_khurana\}@sfu.ca}}

\nipsfinalcopy % Uncomment for camera-ready version

\begin{document}

\maketitle

\begin{abstract}
We address the problem of language-based temporal localization of moments in untrimmed videos. Compared to temporal localization with fixed categories, this problem is more challenging as the language-based queries have no predefined activity classes and may also contain complex descriptions. Current state-of-the-art model MAC\cite{Ge2018MACMA} addresses it by mining activity concepts from both video and language modalities. This method encodes the semantic activity concepts from the verb/object pair in a language query and leverages visual activity concepts from video activity classification prediction scores. We propose ``Multi-faceted Video Moment Localizer'' (MML), an extension of MAC model by the introduction of visual object evidence via object segmentation masks and video understanding features via video captioning. Furthermore, we improve language modelling in sentence embedding. We experimented on Charades-STA dataset and identified that MML outperforms MAC baseline by 4.93\% and 1.70\% on R@1 and R@5 metrics respectively. Our code and pre-trained model are publicly available at \textit{\url{https://github.com/madhawav/MML}}.
\end{abstract}

\section{Introduction}\label{sec:intro}
Imagine being able to search for the moment in a video where an adorable kitten sneezes, even though the uploader has not tagged its timestamp. Finding such important events in a video is called a temporal moment localization. Videos contain many untagged topics and hence, addressing temporal localization problem is versatile. For instance, progress on temporal moment localization can benefit video search, video summarization, action moment detection and many other areas.

\begin{figure}[t]
    \centering
    \includegraphics[width=\linewidth]{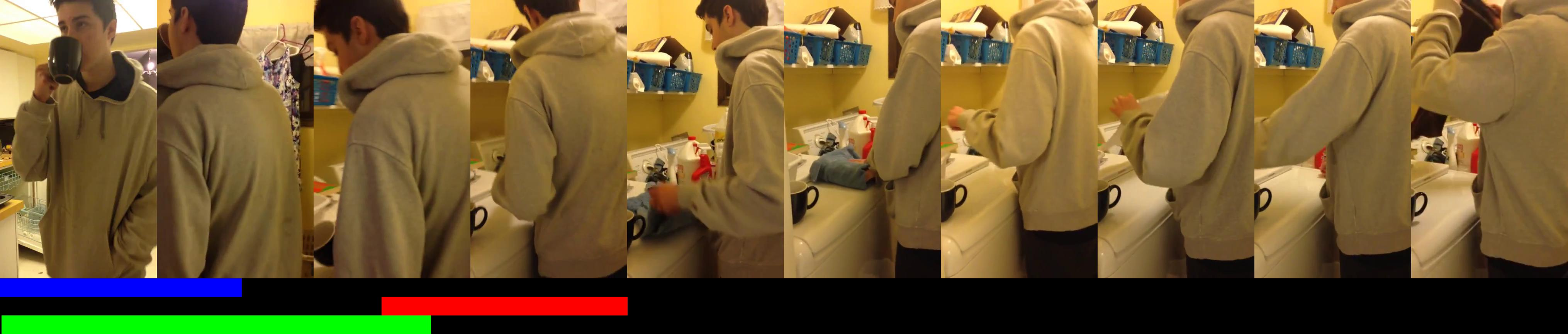}
  \caption{An example for query \textit{``Person drinking a cup of coffee''}. Green, red and blue lines indicate the ground truth frames, baseline~\cite{Ge2018MACMA} prediction and prediction by our method respectively.}
    \label{fig:example-coffee}
\end{figure}

We address the problem of language based  action localization in untrimmed videos, where the task is to identify the temporal location within a video that is described by a given natural language query. For example in Figure~\ref{fig:example-coffee}, if we were to query \textit{``Person drinking a cup of coffee''}, the network should locate frames in the video where a person is drinking a coffee. This problem is challenging due to the complexity of action identification in an untrimmed video that may contain a diverse combination of actors, actions and objects over time.

A prior work, Temporal Activity Localization via Language (TALL) \cite{Gao2017TALLTA} compares visual features with sentence embeddings via a Multi-Modal Processing Unit (MPU). It maps the visual and textual features to a single feature space by performing element-wise addition, element-wise multiplication and concatenation. Then, the concatenated output of these operations is used to generate a visual-semantic alignment score along with a location regression result for each video clip. However, visual features used by \cite{Gao2017TALLTA} are trained for activity classification and not to compare with text.

Recent work by Ge et al.~\cite{Ge2018MACMA}, argues that TALL~\cite{Gao2017TALLTA} model ignores rich semantic information about activities in videos and queries. They improved TALL architecture and proposed Mining Activity Concepts (MAC), which extracts activity concepts from verb/object pairs of query sentence and videos. The verb/object pair represents semantic activity concepts that identify the action of a query. The video features from the last layer (FC8) of a C3D model~\cite{Tran2014C3DGF}, trained on Kinetics activity classification dataset, represents visual activity concepts. These activity concepts can be combined using a MPU to predict video moments. Although some Kinetics classes identify an object (e.g. playing guitar), it ignores a large number of object classes contained in a video. 

To overcome above mentioned limitations, we propose Multi-Faceted Moment Localizer (MML) that uses object features from semantic segmentation model and video understanding features from a video captioning model. Our contributions can be summarized as follows:
\begin{itemize}
    \item Introducing frame-level object evidence via semantic object segmentation features to explicitly identify the relationship between the query object and visible objects in video frames.
    \item Introducing video captioning features as an additional feature for joint-embedding between video clip and query text.
    \item Improved language modelling of query text via the introduction of BERT sentence embedding. Experiments on improving Verb/Object pair (VO pair) encoding via BERT which did not improve results, and we provide the reasons that may have caused this counter-intuitive outcome in Section~\ref{subsec:bert}.
\end{itemize}
We performed extensive ablation study to validate each architectural component we introduced. Our code and pre-trained model is available at \textit{\url{https://github.com/madhawav/MML}}.

\section{Approach}\label{sec:approach}
\begin{figure}[t]
\centering
\includegraphics[scale=0.19]{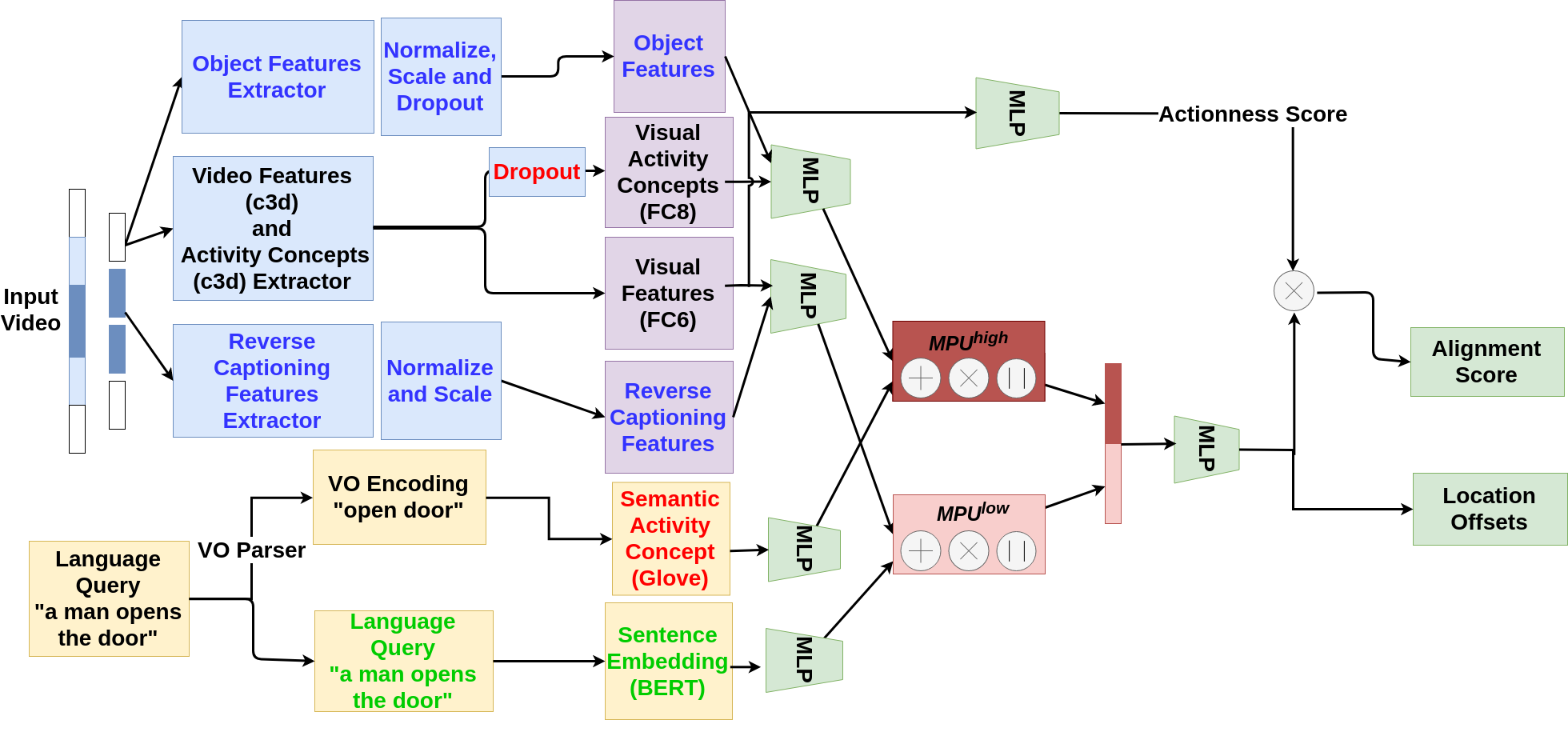}
\caption{Architecture diagram of MML. Blocks with blue text indicate components we introduced. Blocks with green text indicate components we improved. Blocks with red text indicate components considered for improvements. Blocks with black text indicate other components from baseline~\cite{Ge2018MACMA}.}
\label{fig:archdiag}
\end{figure}

The proposed work Multi-Faceted Moment Localizer (MML) is based on the prior work MAC \cite{Ge2018MACMA}, which is the current state-of-the-art method for text based video moment localization. We developed our model on top of a PyTorch implementation of MAC available here~\footnote{https://github.com/WuJie1010/Temporally-language-grounding}. We focused on improving the MAC architecture by introducing additional features that support video moment localization. Figure~\ref{fig:archdiag} shows the architecture of our model in which the blocks with green and blue colour text indicate our contributions.

Similar to the approach taken in MAC, we first divide a video into overlapping video clips. Then an alignment score and location offset pair is calculated considering the query sentence and each video clip. This calculation involves: 1) comparison of low-level video clip features (C3D FC6 features from MAC and video captioning features we introduced) with sentence embedding, 2) comparison of high-level video clip features (visual activity concepts from MAC and object segmentation features we introduced) with VO pair glove embedding.

Two Multi Modal Processing Units \cite{Gao2017TALLTA} ($\text{MPU}^{\text{low}}$ and $\text{MPU}^{\text{high}}$) are used for above mentioned comparisons. The outputs from two MPU's are concatenated and passed through an additional fully connected layer (MLP) to obtain alignment score and location offsets. Following the same approach as MAC, the alignment score is then multiplied by the actionness score, which indicates the likelihood of the candidate video clip to have meaningful activities. The location offsets are added to video clip time bounds to obtain a particular prediction.

\subsection{Contribution 1: Improvements on language modelling}\label{subsec:bert}
Both TALL~\cite{Gao2017TALLTA} and MAC~\cite{Ge2018MACMA} models use skip-thought vectors for sentence embedding.
In our approach, we replaced skip-thought embeddings  using Google’s pre-trained BERT~\cite{devlin2018bert} model features.
The intuition behind this change is two-fold: 1) Skip-thoughts is a 4800-dimensional vector whereas BERT is a 768-dimensional vector, thereby reducing the parameter count and improving the generalization of the model; and 2)  BERT is trained on a significantly larger datasets: BooksCorpus (800M words) and Wikipedia (2,500M words).
As explained in Section~\ref{sec:experiments}, the introduction of BERT sentence embeddings improved the results of our model.  We also tried 
Facebook AI’s RoBERTa~\cite{liu2019roberta}, a derivative of BERT.
But RoBERTa reduced the performance of our model. This reduction is probably because RoBERTa has removed the next sentence prediction training objective, which had made BERT versatile for sentence embedding.

The MAC~\cite{Ge2018MACMA} model uses GloVe embedding for VO word pairs. We tried using BERT as a substitute for GloVe, but that only degraded results. This is probably because BERT embedding (768d) is much larger than GloVe embedding (300d), which causes the number of parameters in the model to increase, and it may have influenced the generalization of the model.

\subsection{Contribution 2: Introducing object segmentation features}\label{sec:intro-object-segment}
The baseline~\cite{Ge2018MACMA} uses FC8 features (i.e. action class predictions) of a C3D model trained on Kinetics dataset as visual activity concepts used to identify actions in a clip. However, only 15.82\% of the objects mentioned in the query text of Charades-STA dataset are covered by Kinetics classes. To overcome this issue, we used object segmentation features to explicitly include object information contained in clip.

\begin{figure}[t]
\centering
\includegraphics[scale=0.4]{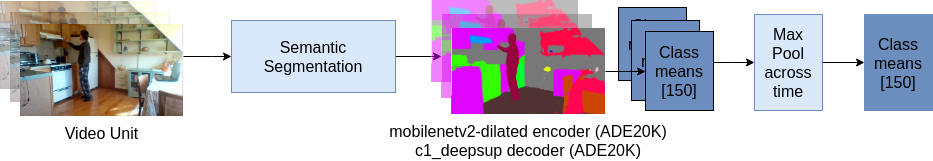}
\caption{Process of extracting object segmentation features.}
\label{fig:obj-feature-extracts}
\end{figure}

For identifying objects in video frames, we considered semantic segmentation models trained on ADE20K dataset~\cite{ADE20K} which consists of 150 object classes. From the provided pre-trained object segmentation models \footnote{https://github.com/CSAILVision/semantic-segmentation-pytorch}, we used \textit{``MobileNetV2dilated + C1\_deepsup''} in the interest of time, which provided required features in about 5 days. We sampled every $16^{th}$ frame of video clips and obtained each frame's class distribution using the semantic segmentation model. We obtained the class means of each frame and then max pooled across the time dimension. The above process resulted in a 150 dimension vector ($V_{\text{obj}}$) for each video clip. Figure~\ref{fig:obj-feature-extracts} demonstrates the aforementioned object feature extraction process. 
At training time, $V_{\text{obj}}$ is normalized, scaled (using scale ratio $S_{\text{obj}} \in [0,1]$) and concatenated with Visual Activity Concept features ($V_{\text{vac}}$), before being passed to the MLP that generate the input for $\text{MPU}^{\text{high}}$. In order to address over-fitting, dropout layers (with dropout ratios $D_{\text{obj}}$ and $D_{\text{vac}}$) were used for $V_{\text{obj}}$ and $V_{\text{vac}}$. The input to the MLP is as follows:
\[
MLP_{input} = \textbf{Dropout}(\frac{V_{\text{obj}}}{|V_{\text{obj}}|} \cdot S_{\text{obj}}, \text{dropout\_ratio}=D_{\text{obj}}) \mathbin\Vert \textbf{Dropout}(V_{\text{vac}},\text{dropout\_ratio}=D_{\text{vac}}) 
\]
In order to identify suitable hyper-parameter values for $S_{\text{obj}}$, $D_{\text{obj}}$ and $D_{\text{vac}}$, we used a 3 axis parameter sweep as explained in Section~\ref{sec:param-sweep}.

\subsection{Contribution 3: Introducing video captioning features}
Existing methods for video moment localization use C3D features from activity classification domain. However, the task of moment localization is different from activity classification as we have to compare data from two different domains: highly complex video domain and a relatively simple text domain. Thus, it may be helpful to use features from a model that compares videos with text, and thus we incorporate video captioning features into our model.

We use Temporal Shift Module (TSM)~\cite{lin2019tsm} for this task, as it is the state of the art method in video understanding. We have used \textit{TSM ResNet50}~\footnote{https://github.com/mit-han-lab/temporal-shift-module} model pre-trained on $16$ frames to extract video captioning features. This model was trained on Kinetics-400 dataset. Hence, it outputs features per frame. We perform average pooling across all frames of a clip to obtain one feature vector of size $2048$ for each video clip. These features denote a low-level representation of the video clip. Hence, we concatenate these features with FC6 features of C3D model to calculate the alignment score and frame offset. Result of using video captioning features is given in Table~\ref{table:quantitative-feature-selection}.

\section{Experiments} \label{sec:experiments}

\subsection{Dataset}
We used Charades-STA~\cite{Gao2017TALLTA} dataset for training and evaluating the MML model. This dataset contains around 10,000 videos where each video contains clip-level sentence descriptions coupled with start/end time-stamps. In total, there are 13898 clip-sentence pairs in Charades-STA training set and 4233 clip-sentence pairs in test set. To keep our results comparable, we used the same train/test splits of Charades STA as baseline\cite{Ge2018MACMA} for all the experiments.

\subsection{Evaluation Metrics}
We adopted the evaluation metrics \textit{R@1 on IOU=0.5} and \textit{R@5 on IOU=0.5} used by baseline\cite{Ge2018MACMA}. \textit{R@N on IOU=u} can be calculated as $\frac{1}{N_q}\sum_{i=1}^{N_q} r(n, u, q_i)$ where $r(n,u,q_i)$ is the alignment result for query $q_i$. Here, $r(n,u,q_i) = 1$ indicates correct alignment and $r(n,u,q_i) = 0$ indicates wrong alignment, considering the top $n$ scored video clips having a temporal IOU greater than or equal to $u$ with the ground truth of $q_i$. The value $N_q$ represents the total number of queries and thus \textit{R@n on IOU=u} is an averaged figure. A higher value is preferred for\textit{ R@n on IOU=u}. All measurements in this paper are at \textit{IOU = 0.5}.

\subsection{Baseline}
The baseline model by MAC \cite{Ge2018MACMA} authors exhibit \textit{R@1} of $0.304$ and \textit{R@5} of $0.648$ on Charades-STA dataset. However, the PyTorch implementation~\footnote{https://github.com/WuJie1010/Temporally-language-grounding} of the same model provide  slightly different values at $0.297$ for R@1 and $0.641$ for R@5. The MML was developed from the code base of this PyTorch implementation. Thus, we consider both of these as baselines for our experiments.

\subsection{Experimental results}
The best models we identified provide 4.93\% (R@1) and 1.70\% (R@5) improvements over MAC author's baseline. In this subsection, we first discuss the effect of each component of the architecture using an ablation study. Then we provide details of hyper-parameter tuning of object segmentation features and video captioning features, followed by a qualitative analysis of the proposed model.

\begin{table}[t]
\caption{Effect of different features on performance of MML. R@1 and R@5 are used as evaluation metric. Best performing models are highlighted in blue color.}
\label{table:quantitative-feature-selection}
\centering
\resizebox{\textwidth}{!}{%
\begin{tabular}{@{}l|llll|ll@{}}
\toprule
\multicolumn{1}{r|}{} &  &  &  &  &  &  \\
\multicolumn{1}{r|}{\multirow{-2}{*}{\textbf{Features}}} &  &  &  &  &  &  \\
\textbf{Models} & \multirow{-3}{*}{\begin{tabular}[c]{@{}l@{}}\textbf{Sentence}\\ \textbf{Embedding}\end{tabular}} & \multirow{-3}{*}{\begin{tabular}[c]{@{}l@{}}\textbf{VO} \\ \textbf{Embedding}\end{tabular}} & \multirow{-3}{*}{\begin{tabular}[c]{@{}l@{}}\textbf{Object}\\ \textbf{Segmentation}\\ \textbf{Features}\end{tabular}} & \multirow{-3}{*}{\begin{tabular}[c]{@{}l@{}}\textbf{Video}\\ \textbf{Captioning}\\ \textbf{Features}\end{tabular}} & \multirow{-3}{*}{\textbf{R@1}} & \multirow{-3}{*}{\textbf{R@5}} \\ \midrule
\begin{tabular}[c]{@{}l@{}}\textbf{MAC} \\ \textbf{(authors baseline)}\end{tabular} & SkipThought & GloVe & - & - & 0.304 & 0.648 \\
\begin{tabular}[c]{@{}l@{}}\textbf{MAC}\\ \textbf{(PyTorch baseline)}\end{tabular} & SkipThought & GloVe & - & - & 0.297 & 0.641 \\
\textbf{Model 1} & BERT & GloVe & - & - & 0.299 & 0.647 \\
\textbf{Model 2} & SkipThought & GloVe & \checkmark & - & 0.308 & 0.646 \\
{\color[HTML]{3166FF} \textbf{Model 3}} & {\color[HTML]{3166FF} \textbf{BERT}} & {\color[HTML]{3166FF} \textbf{GloVe}} & {\color[HTML]{3166FF} \textbf{\checkmark}} & {\color[HTML]{3166FF} -} & {\color[HTML]{3166FF} \textbf{0.313}} & {\color[HTML]{3166FF} \textbf{0.659}} \\
\textbf{Model 4}& SkipThought & BERT & \checkmark & - & 0.302 & 0.642 \\
\textbf{Model 5} & BERT & BERT & \checkmark & - & 0.301 & 0.647 \\
\textbf{Model 6}& RoBERTa & GloVe & \checkmark & - & 0.238 & 0.574 \\
{\color[HTML]{3166FF} \textbf{Model 7}} & {\color[HTML]{3166FF} \textbf{BERT}} & {\color[HTML]{3166FF} \textbf{GloVe}} & {\color[HTML]{3166FF} \textbf{\checkmark}} & {\color[HTML]{3166FF} \textbf{\checkmark}} & {\color[HTML]{3166FF} \textbf{0.319}} & {\color[HTML]{3166FF} \textbf{0.651}} \\ \bottomrule
\end{tabular}%
}
\end{table}

\subsubsection{Ablation study on feature selection}
We perform extensive ablation study to validate the effectiveness of individual components of the proposed model. 
Table \ref{table:quantitative-feature-selection} illustrates R@1 and R@5 scores of various ablations studies. Introduction of BERT sentence embedding and object segmentation features help outperform the baseline. The best models we identified (Model 3 and Model 7) necessarily includes BERT sentence embedding, object segmentation features and GloVe VO embeddings. Introduction of video captioning features further improves R@1. These best models provide 4.93\% (R@1 of Model 7) and 1.70\% (R@5 of Model 3) improvements over MAC authors baseline. Considering the model 3 alone, we obtain 2.96\% (R@1) and 1.70\% (R@5) improvements.

The models 3 and 7, which are the best performing MML models, achieve 7.41\% and 4.93\% improvements in R@1 over MAC PyTorch baseline. The introduction of object segmentation features (model 2 vs. baseline) and video captioning features (model 7 vs. model 2) boosts R@1 by 3.70\% and 2.02\% respectively. Although BERT sentence embedding alone contribute only a 0.67\% increase in R@1 score, its combination with object segmentation features provide a further improvement of 1.01\%.

\subsubsection{Hyper-parameter tuning on object segmentation and video captioning features}\label{sec:param-sweep}
As mentioned in Section \ref{sec:intro-object-segment}, we consider multiple values for hyper-parameters $D_{\text{obj}}$, $D_{\text{vac}}$ and $S_{\text{obj}}$ to address over-fitting. We used a 3 axis parameter sweep to find a suitable set of hyper-parameters considering candidate values $S_{\text{obj}} \in \{ 0,0.005,0.05,0.1,0.25,0.5,0.75,1\}$, $D_{\text{obj}} \in \{0,0.1,0.25,0.5\}$ and $D_{\text{vac}} \in \{0,0.1,0.25,0.5\}$. We effectively validated $8\times4\times4=128$ models that use BERT sentence embedding, GloVe VO embedding and no video captioning features, and identified $D_{\text{obj}}=0.5, D_{\text{vac}}=0$ and $S_{\text{obj}}=0.005$ as the best configuration. This configuration resulted in model 3 in Table \ref{table:quantitative-feature-selection}. The complete parameter sweep took about 2 days to complete on a machine with RTX 2080Ti GPU, having cached all the features in the system memory.

\begin{figure*}[t]
    \centering
    \begin{subfigure}{0.5\textwidth}
        \centering
        \includegraphics[scale=0.18]{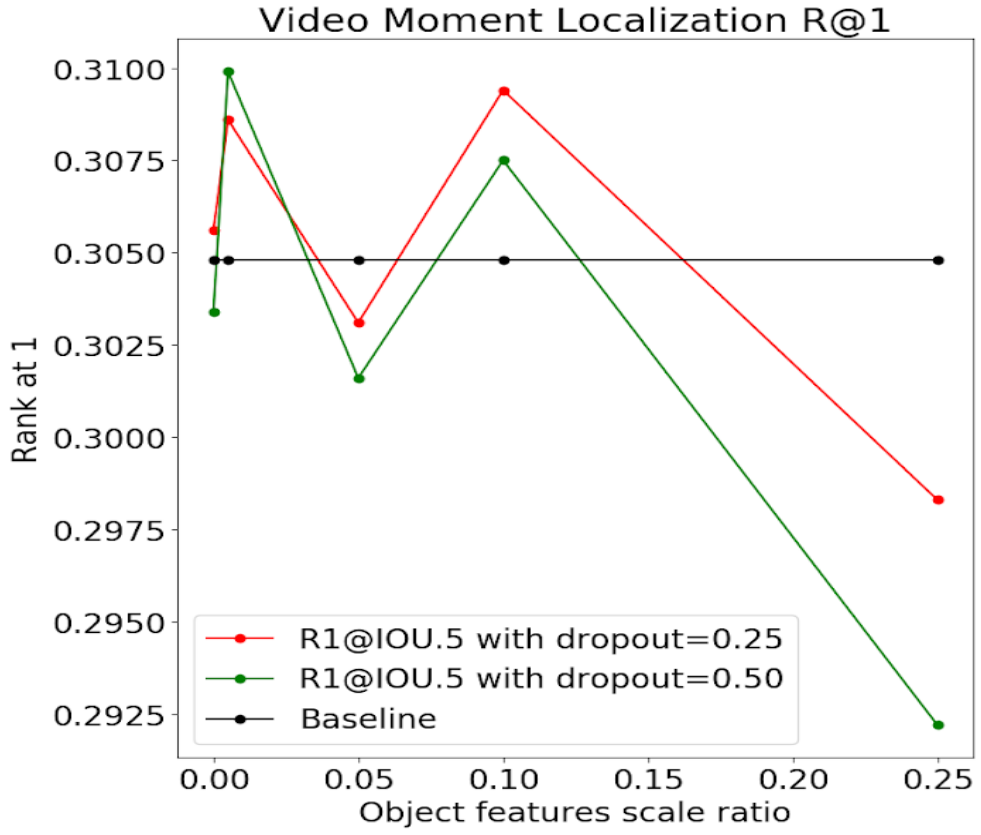}
        \caption{R@1 (IOU=0.5) on test set}
    \end{subfigure}%
    ~ 
    \begin{subfigure}{0.5\textwidth}
        \centering
        \includegraphics[scale=0.18]{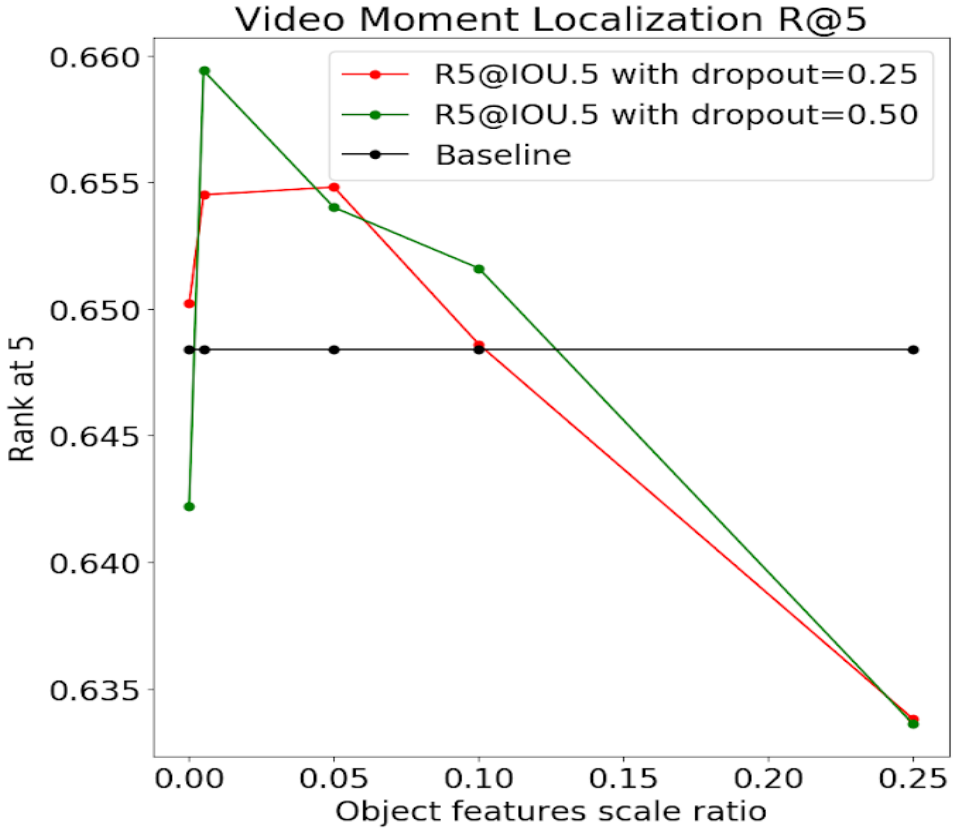}
        \caption{R@5 (IOU=0.5) on test set}
    \end{subfigure}
    \caption{Metrics at various $S_{\text{obj}}$ and $D_{\text{obj}}$. $D_{\text{vac}}$ is kept at 0. BERT sentence embedding, GloVe VO embedding and no video captioning features are used. Baseline by MAC authors is provided.}
    \label{fig:obj-param-tune}
\end{figure*}

Figure~\ref{fig:obj-param-tune} shows the plots of R@1 and R@5 of best models at various scale ratios and dropout ratios of object features. The best models are identified by validating after each epoch. $D_{\text{vac}}$ is set to 0 (no visual activity concept dropout) as it is identified to be the best configuration. A significant reduction in R@1 and R@5 is observed when $S_{\text{obj}}$ is increased, indicating the models as over-fitting.

We performed a similar hyper-parameter search for finding a suitable scale for incorporating video captioning features, based on the best performing hyper-parameters for object segmentation features. We obtained the best results with a scale of $0.005$. The results are shown in model 7 of Table~\ref{table:quantitative-feature-selection}.

\subsubsection{Qualitative results}
\begin{figure}[h!]
\centering
\begin{subfigure}{.49\textwidth}
  \centering
  \includegraphics[scale=0.028]{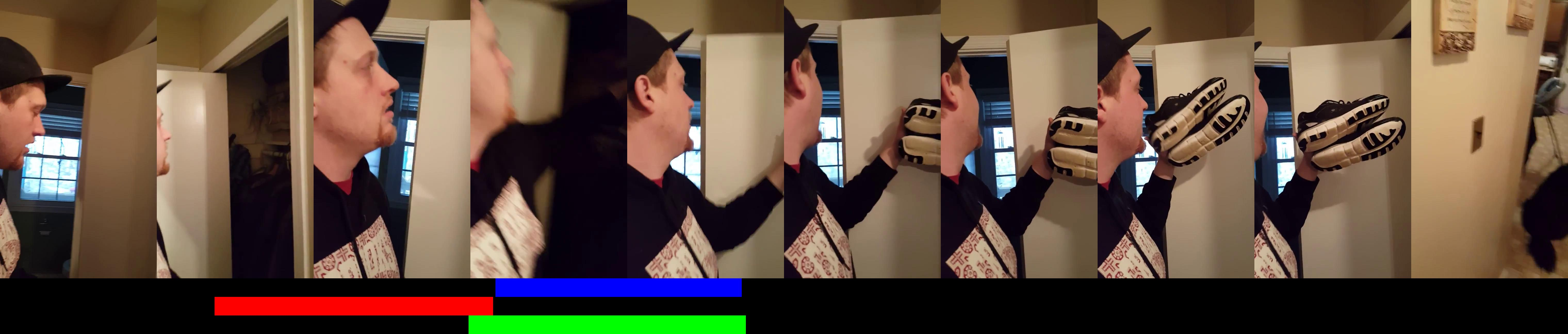}
  \caption{person hold the shoes}
  \label{fig:qualitative-shoe}
\end{subfigure}
\begin{subfigure}{.49\textwidth}
\centering
  \includegraphics[scale=0.040]{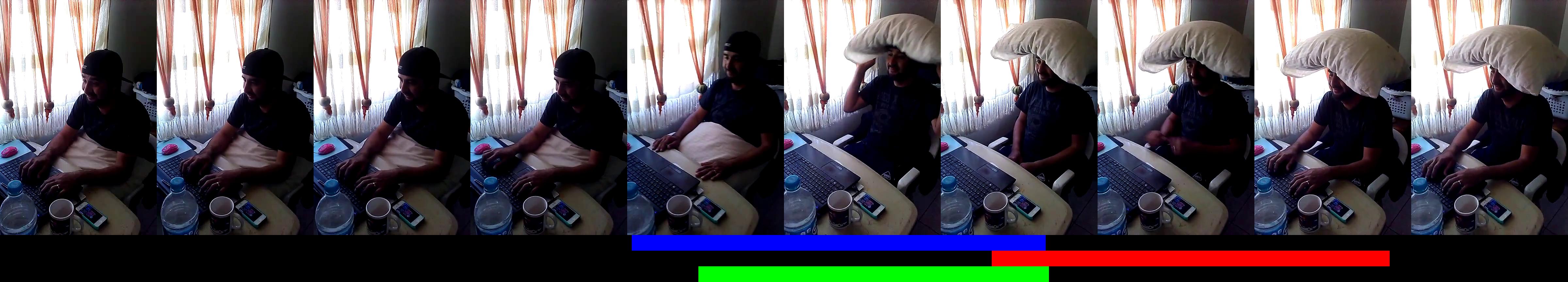}
  \caption{person puts a pillow on their head}
      \label{fig:qualitative-pillow}
\end{subfigure}
\begin{subfigure}{.49\textwidth}
   \centering
  \includegraphics[scale=0.032]{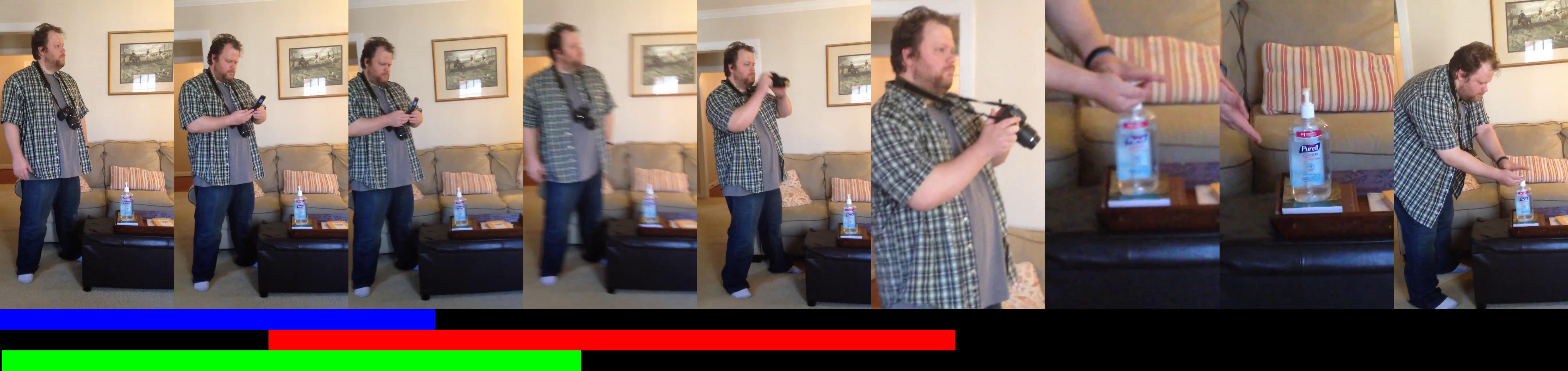}
   \caption{the person is playing with a phone}
          \label{fig:qualitative-phone}
\end{subfigure}
\begin{subfigure}{.49\textwidth}
   \centering
  \includegraphics[scale=0.032]{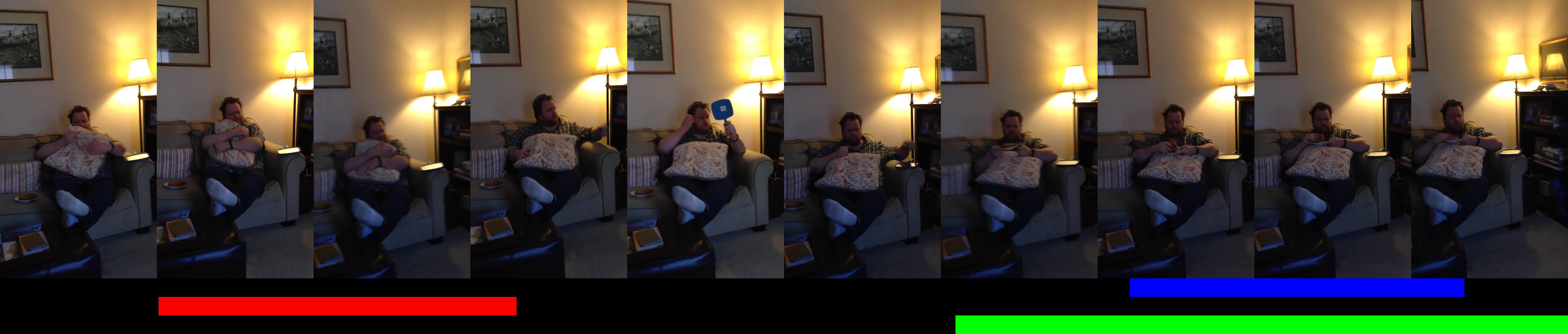}
   \caption{person finally eating a sandwich in a living room}
      \label{fig:qualitative-sandwich}
\end{subfigure}
\begin{subfigure}{.49\textwidth}
  \centering
  \includegraphics[scale=0.031]{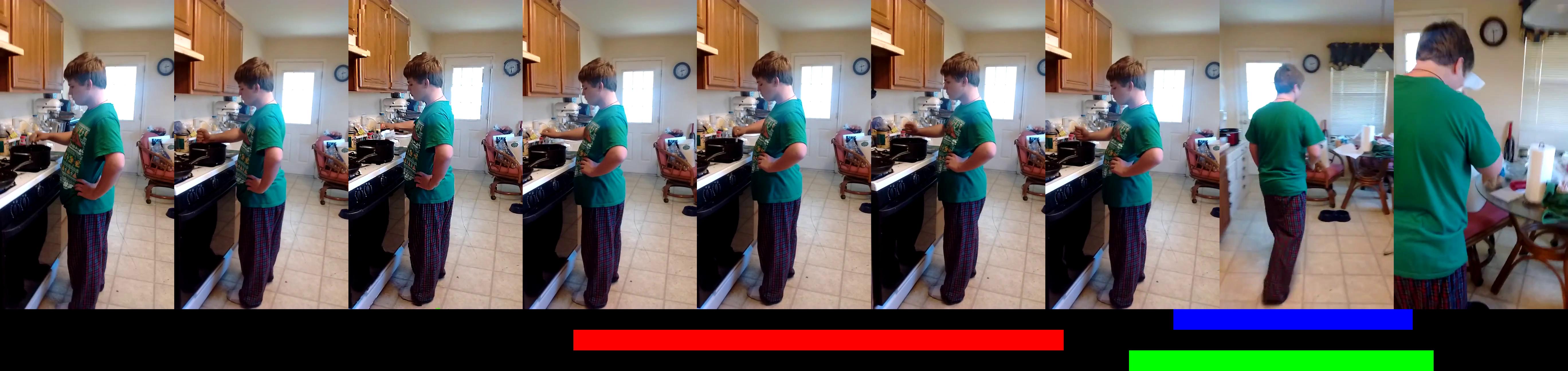}
   \caption{the person takes a paper towel from the table}
           \label{fig:qualitative-paper-towel}
\end{subfigure}
\begin{subfigure}{.49\textwidth}
   \centering
  \includegraphics[scale=0.048]{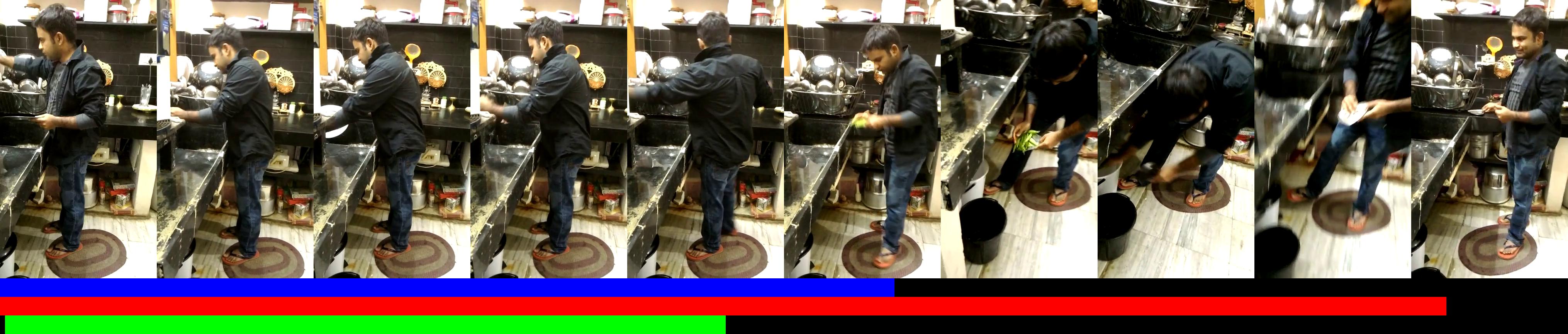}
   \caption{the person washes dishes over the sink}
        \label{fig:qualitative-dishes}
\end{subfigure}
\caption{Selected top-1 results with improvements. Green lines, red lines and blue lines indicate ground truths, MAC results and MML results respectively. Queries are provided below each figure.}
\label{fig:qualitative-study-positive}
\end{figure}

\begin{figure}[h!]
\centering
% \par\bigskip % force a bit of vertical whitespace
\begin{subfigure}{.39\textwidth}
\centering
  \includegraphics[scale=0.025]{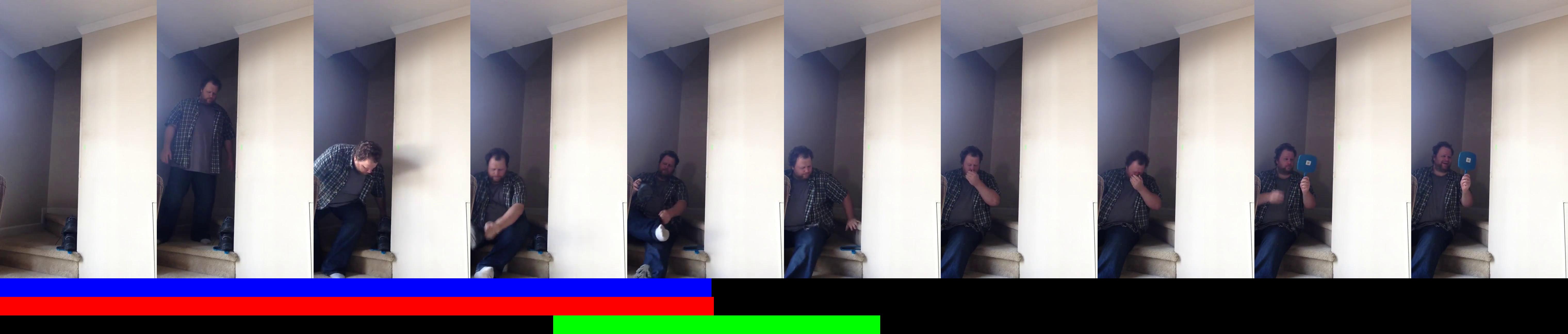}
   \caption{person puts shoes on}
   \label{fig:qualitative-shoes-on}
\end{subfigure}
\begin{subfigure}{.49\textwidth}
  \centering
  \includegraphics[scale=0.035]{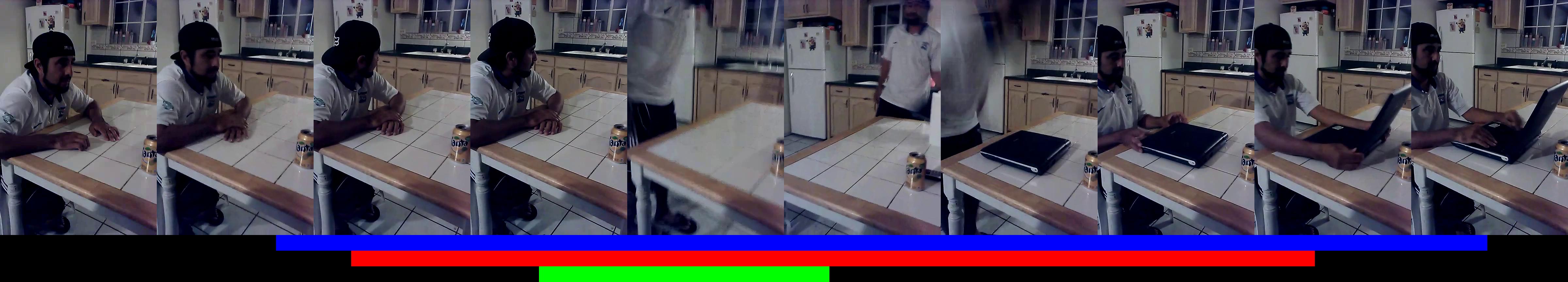}
  \caption{the person stands up}
  \label{fig:qualitative-person-stand-up}
\end{subfigure}
\caption{Selected top-1 results with no improvements. Green lines, red lines and blue lines indicate ground truths, MAC results and MML results respectively. Queries are provided below each figure.}
\label{fig:qualitative-study-no-change}
\end{figure}

Figures \ref{fig:qualitative-study-positive} and \ref{fig:qualitative-study-no-change} show qualitative examples from MML compared to the MAC baseline. These examples are from Model 7 in Table \ref{table:quantitative-feature-selection}. Below each example, ground truth (green stripe), MAC prediction (red stripe) and MML prediction (blue stripe) are indicated.

Figure \ref{fig:qualitative-study-positive} shows examples where the use of MML improved results over the MAC baseline. The improvement on examples \ref{fig:qualitative-shoe}, \ref{fig:qualitative-pillow}, \ref{fig:qualitative-phone}, \ref{fig:qualitative-sandwich} and \ref{fig:qualitative-paper-towel} can be largely attributed to the introduction of object segmentation features. This is because, in these examples, the objects referred by the queries enter/leave visibility at the vicinity of predictions. 
In the example \ref{fig:qualitative-dishes}, improvements are made despite the object (``shoe'') being on the scene throughout the video. Thus, BERT sentence embedding and video captioning features may have played a role in these improvements.

Figure \ref{fig:qualitative-study-no-change} shows two examples where improvements of MML model did not improve the results. In Figure \ref{fig:qualitative-shoes-on}, the object does not change visibility and therefore, the results are identical to the baseline. Figure \ref{fig:qualitative-person-stand-up} shows an example where MML has performed worse than MAC baseline. In this case, there is no object in the sentence.


\section{Conclusion}\label{sec:conclusion}
In this project, we addressed the problem of video moment localization and proposed a novel method \textit{Multi-faceted Video Moment Localization} (MML). Our model is built on top of the current state-of-the-art model MAC~\cite{Ge2018MACMA}. We introduced BERT sentence features for the text query, and object segmentation features and video captioning features for the video, thereby improving language based localization of moments in a given video. We performed an extensive ablation study to validate the effectiveness of each component introduced in MML. Our experiments show the significance of our proposed method over the current baselines.

\bibliographystyle{unsrt}       % APS-like style for physics
\bibliography{main}   % name your BibTeX data base

\end{document}